\documentclass[
 aapm,
 mph,
 amsmath,
 amssymb,
 reprint
]{revtex4-1}

\usepackage{graphicx}
\usepackage{dcolumn}
\usepackage{bm}

\usepackage[mathlines]{lineno}
\usepackage{natbib}
\bibpunct[:]{(}{)}{;}{a}{,}{,}

\begin{document}
\title{Kernel-based framework to estimate deformations of pneumothorax lung \\using relative position of anatomical landmarks}

\author{Utako Yamamoto$^a$}
 \email{Corresponding author.}
\author{Megumi Nakao$^a$}
\author{Masayuki Ohzeki$^{bc}$}
\author{\\Junko Tokuno$^d$}
\author{Toyofumi Fengshi Chen-Yoshikawa$^e$}
\author{Tetsuya Matsuda$^a$}

\email{E-mail addresses: yamamoto.utako.6a@kyoto-u.jp (U. Yamamoto)}

\affiliation{\footnotesize 
$^a$Department of Systems Science, Graduate School of Informatics, Kyoto University,  Kyoto, Japan,}

\affiliation{\footnotesize 
$^b$Graduate School of Information Sciences, Tohoku University, Sendai, Japan,}

\affiliation{\footnotesize 
$^c$Institute of Innovative Research, Tokyo
Institute of Technology, Yokohama, Japan,}

\affiliation{\footnotesize 
$^d$Department of Thoracic Surgery, Graduate School of Medicine, Kyoto University, Kyoto,  Japan,}

\affiliation{\footnotesize 
$^e$Department of Thoracic Surgery, Graduate School of Medicine, Nagoya University, Nagoya, Japan}

\begin{abstract}
\noindent
In video-assisted thoracoscopic surgeries, 
successful procedures of nodule resection are highly dependent on 
the precise estimation of lung deformation between the inflated lung in the computed tomography 
(CT) images during preoperative planning 
and the deflated lung in the treatment views during surgery. 
Lungs in the pneumothorax state during surgery have a large volume change from normal lungs, making it difficult to build a mechanical model.
The purpose of this study is to develop a deformation estimation method of 3D surface of a deflated lung from a few partial observations. 
To estimate deformations for a largely deformed lung, a kernel regression-based solution was introduced. 
The proposed method used a few landmarks to capture the partial deformation between the 3D surface mesh obtained from preoperative CT and the intraoperative anatomical positions. 
The deformation for each vertex of the entire mesh model was estimated per-vertex as a relative position from the landmarks. 
The landmarks were placed in the anatomical position of the lung's outer contour.
The method was applied on nine datasets of the left lungs of live beagle dogs. 
Contrast-enhanced CT images of the lungs 
were acquired.
The proposed method achieved a local positional error of vertices of 2.74 mm, Hausdorff distance of 6.11 mm, and Dice similarity coefficient of 0.94.
Moreover, the proposed method achieved the estimation lung deformations from a small number of training cases and a small observation area.
This study contributes to data-driven modeling of pneumothorax deformation of the lung.


\end{abstract}

\maketitle
\section{\label{sec:Introduction}Introduction}

Medical imaging technology, such as computed tomography (CT), has become common in clinical examination. 
The CT screening can detect lung nodules at a small size and early stage \citep{Swensen2005}, 
and such early detection has been shown to increase survival rate \citep{Mikita2012}. 
Video-assisted thoracoscopic surgery \citep{Shaw2008,Flores2008} is widely performed as a minimally invasive surgical technique. 
However, the minimally invasive approach can complicate localizing small nodules during surgery. 
The inflated lung (at inhale or exhale) in preoperative CT images 
turns collapsed to a fully deflated state during surgery, creating a large lateral pneumothorax. 
Although lung nodules are detected by preoperative CT images during preoperative surgical planning, 
the position of nodules may change because of the gross deformation of the lung induced by this process. 
Various adjunctive techniques, such as nodule tagging by physical or chemical markers \citep{Sato2014,Lin2016}, 
are used to localize nodules during surgery. 
Such localization is carried out preoperatively, 
requiring additional CT imaging and invasive tagging procedure. 
Therefore, it increases the clinical burden on both surgeons and patients. 
Instead, a less invasive technique that can estimate the intraoperative position of the lung nodules accurately is highly desirable 
for more precise nodule resection and better preservation of normal lung function 
as a consequence of minimizing resection margins. 

Accurate estimation of whole lung deformation induced by a large pneumothorax 
is required to estimate the intraoperative position of nodules 
by evaluating the correspondence of the lung surface and the internal structure with nodules. 
Deformation estimation for the soft tissue of the lung has been explored 
as a deformable image registration problem between CT images \citep{Ruhaak2017,Sotiras2013}. 
In deformable image registration in thoracic contexts, 
respiratory motion \citep{Wilms2016,Rietzel2006,Yin2011} and patient posture \citep{Nakamoto2007} 
have been the main focus of investigation. 
For the analysis of respiratory motion, 
image-based lung modeling techniques \citep{Fuerst2015,Nakao2007} and 
statistical modeling techniques \citep{Ehrhardt2011,Jud2017} have been developed previously. 

Unlike respiratory motion, 
lung deformation by pneumothorax can induce considerable volume change. The mechanism is complex and is not mathematically understood. 
To compute an extrapolation to the deflated lung, 
the displacement field obtained from the registration of 
inhale-exhale states in 4D CT was used in \cite{Naini2009,Naini2011}.
In the field of intraoperative registration of the collapsed lung, 
a registration framework to analyze the displacement of 
internal lung structures on cone-beam CT (CBCT) using animal lungs \citep{Uneri2013} 
and a deformable registration method for postural differences 
on preoperative CT and intraoperative CBCT \citep{Chabanas2018} 
have been proposed. 
The CBCT imaging provides invaluable real-time information to the surgeons 
within the small acquisition time. 
Those researches on deformation registration using CBCT 
has focused mainly on the registration of internal structures.

To provide subsurface nodule position from the lung surface information during surgery, 
it is required to register the preoperative CT image to intraoperative physical lung 
by the surface information of the lung. 
Intraoperative segmentation of thoracoscopic camera images \citep{Wu2017} 
and registration methods using the 2D appearance or silhouettes of 
intraoperative collapsed lung surface as visual cues 
to register preoperative CT models with intraoperative camera images \citep{Saito2015,Nakao2017} 
have been reported. 
Although they achieved the registration from little information 
that can be obtained intraoperatively, 
they had a limitation on the registration of depth direction. 
\cite{Nakamoto2006} proposed intraoperative registration methods 
using several surface points measured on the deflated lungs. 

The deformation estimation problem is a framework 
that provides more localized deformation of each position in the whole shape
than that of the deformable image registration problem. 
The finite element method (FEM) \citep{Belytschko2014} is commonly used 
as a technique for calculating 
the deformation of a living body, based on a mechanical model.
Using preoperative CT, 
attempts to simulate and visualize operations, such as grasping or 
incising organs during surgical operations, have been reported \citep{Nakao2010}.
Deformation estimation techniques for the shape matching 
based on a dynamic deformation model have been proposed 
for surgical assistance \citep{Suwelack2014}.
However, it is difficult to apply FEM for the pneumothorax deformation of the lung, 
because the deformation induces a considerable volume change, 
and it is also difficult to incorporate unknown material parameters 
or uncertain boundary conditions that describe the pneumothorax state. 
Therefore, deformation estimation technique of whole lung surface 
from the partial surface information without that prior knowledge 
is desirable. 

In this study, 
we propose a data-driven scheme to estimate deformations of whole lung surface 
from the partial surface observation. 
Our method employs a few surface points as landmarks of the intraoperative lung
and preoperative CT images to estimate the deformation of the intraoperative deflated lung in the pneumothorax state.
Some data-driven methods were investigated 
to estimate deformations of the liver and the breast tissue. 
In a studies by \cite{Morooka2010,Morooka2012}, 
a model-based calculation of deformation by FEM was combined with 
a data-driven deformation estimation by a neural network 
to accelerate the calculation by FEM. 
Although those methods have the advantage of not requiring knowledge 
of the mechanical properties of organs to estimate deformations, 
they require to input the direction and magnitude of the force, 
which is difficult to be measured during surgery. 
Some previous studies have attempted to use neural networks 
to estimate the force applied to an organ 
from images of the deformed shape \citep{Aviles2014,Greminger2003} 
requiring observation of the whole object. 
The other data-driven methods
could estimate the deformations of the whole liver surface pulled by a surgical tool 
from partial observations of the liver surface using a neural network \citep{Utako2017NN,Morooka2013}, 
deformations of a liver in respiratory motion \citep{Lorente2017},
and deformations of the breast tissue by two compression plates \citep{Martinez2017}. 
As these methods were data-driven, they required large volumes of data calculated by FEM simulating the deformations. 
That strategy is difficult to apply to the deformation problem of the deflated lung 
because an accurate simulation of the pneumothorax deformation of the lung by FEM 
is difficult. 
To avoid the requirement of accurate simulation for the specific organ, 
Pfeiffer et al. has shown that a convolutional neural network can be trained 
on entirely synthetic data of random organ-like meshes 
to estimate a displacement field inside a liver from the liver surface 
\citep{Pfeiffer2019}.

Among the data-driven estimations, 
the kernel method is more suitable for learning 
with a small number of samples. 
The kernel method contains an infinite number of parameters to represent the data and can be fitted to an optimal solution from a small amount of samples.
It is difficult to collect a large number of training images of the lung volume in the pneumothorax state.
In addition, training the kernel method requires 
only one-time computation of the inverse matrix, 
which is much less computationally intensive 
than training a neural network.
It may be possible to obtain good estimation results 
by constructing a large-scale neural network and 
training it using a supercomputer on a large number of samples. 
However, considering the future clinical applications, 
a method that can be implemented in any computing environment is desired.
In a study of non-rigid image registration of the lung during breathing 
\citep{Jud2017}, 
the authors formulated the statistical motion as kernel and 
integrated it into a parametric non-rigid registration. 
We have considered that a kernel method is available to estimate 
deformations of the liver in \cite{Utako2017Cernel_e}. 
However, that method was designed for a specific patient and evaluated 
on the same liver mesh that was used during the training. 
A deformable registration of CT-CBCT and CBCT-CBCT 
using kernel regression framework has been reported in \cite{Nakao2020}. 
The CBCT has a limited field of view and the images 
that can be acquired are limited to the partial shape of the lung. 
In addition, the lungs are affected by motion of the thoracic cavity, 
even in the pneumothorax state, and may contain motion artifacts, 
especially in the pneumothorax lung, which has low contrast 
(due to low air content), 
making it difficult to acquire stable image features from the deflated lung. 
Therefore, we explore landmark-based estimation methods those are robust to image noises and intraoperative conditions. 

In this study, 
we utilize a kernel method as a data-driven representation 
of the transform mapping of the deformation. 
We focus on landmark-based deformation estimation and 
aim to identify anatomical landmarks that contribute to accurate and stable deformation estimation. 
We propose a method to estimate deformations for new cases 
by setting landmarks in a small observable area during surgery 
and learning data from a small number of cases, and evaluate its performance.
The proposed method was evaluated with a 3D surface mesh constructed from CT images of the left lungs of live beagle dogs 
in inflated and deflated states. 
Moreover, we refer to the estimation performance of the affine transformation 
and thin-plate spline methods, 
which have been reported for deformation of the lung surface 
in the field of intraoperative registration of a collapsed lung 
\citep{Uneri2013,Nakamoto2006}.
The computational complexity of those methods is comparable to ours, 
and those methods can calculate the deformation of the entire lung surface from the information of the configured observation points.

The contributions of this study are as follows: 
1) evaluation of the feasibility of kernel technique 
for estimating deformations of the pneumothorax lung with a large volume change, 
and 
2) analysis of observation rate on the lung surface 
and the data variations due to the number of subject cases.

\section{\label{sec:Methods}Materials and methods}
\subsection{\label{sec:acq}Data acquisition}
The contrast-enhanced CT images of the left lungs of 11 live beagle dogs 
were acquired at two bronchial pressures (14 and 2 cm $\mathrm{H_{2}O}$) 
at the Institute of Laboratory Animals, Kyoto University. 
This study was performed under the regulations 
of the Animal Research Ethics Committee of Kyoto University. 
All CT images were acquired on a 16-row multidetector CT scanner 
(Alexion 16, Toshiba Medical Systems, Tochigi, Japan). 
During the procedure, the dogs were maintained under anesthesia with ketamine, xylazine, and rocuronium and underwent tracheal intubation 
and mechanical ventilation by a ventilator 
(Savina 300, Drager AG \& Co. KGaA, Lübeck, Germany). 
A single trocar hole was first made on the chest wall, 
to let air flow into the pleural cavity. 
Using the ventilator, 
the bronchial pressure was set to 14 cm $\mathrm{H_{2}O}$ 
to obtain images of the fully expanded lungs (inflated state), 
and to 2 cm $\mathrm{H_{2}O}$ 
for imaging of the collapsed lung state (deflated state). 
The doctors set the bronchial pressure to 2 cm $\mathrm{H_{2}O}$ to replicate the deflated lungs in real human surgery.
All dogs were placed in a right lateral (decubitus) position 
on the bed of the CT scanner, 
and the two CT image sets of the inflated and deflated states were acquired 
in that order for each dog. 
For the contrast-enhanced CT, 
10 mL of iopamidol contrast agent was injected 
through a lower extremity peripheral vein. 
Scanning was performed 5 s after the injection of iopamidol.

\begin{figure}[tb]
\includegraphics[width=85mm]{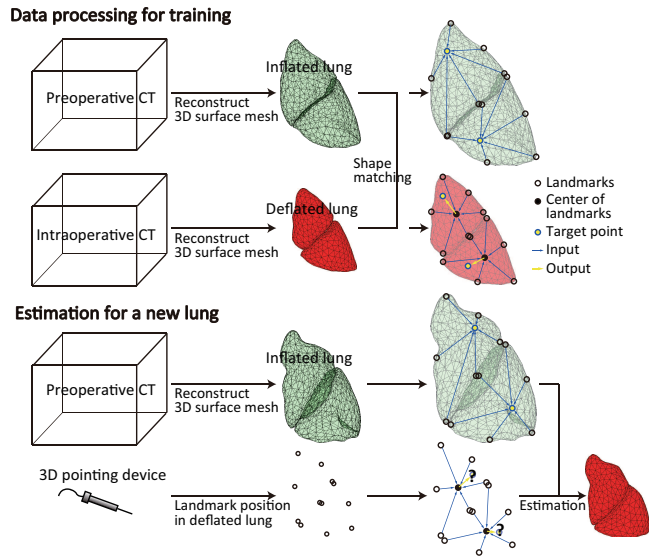}
\caption{Overview of data acquisition, data processing, and input/output for learning and estimation in the deformation estimation framework.}
\label{fig:acq_proc}
\end{figure}

Figure \ref{fig:acq_proc} shows the data acquisition and data processing procedure.
Preoperative and intraoperative CT images were taken in a small number of cases for the use in the regression of known data.
In the CT imaging experiment, the preoperative and intraoperative CT corresponded to bronchial pressures of 14 and 2 cm $\mathrm{H_{2}O}$, respectively.
To estimate the deformations of the lung in new cases, preoperative CT is taken.
Because it is a common procedure to obtain CT images 
of the entire lung prior to surgery, 
preoperative CT data would be available 
when the proposed method is applied to humans. 
Preoperative CT provides data on the inflated lung. 
Intraoperatively, only the positions of the anatomical landmarks on the deflated lung surface are expected to be measured with a 3D pointing device.
In this study, we used CT images that were also taken in a deflated lung 
to obtain the positions of the landmarks.
The setting of the landmarks is described in Section \ref{sec:Landmark_selection}.

\subsection{\label{sec:Data_pre}Data preprocessing}
\begin{figure}[tb]
\includegraphics[width=75mm]{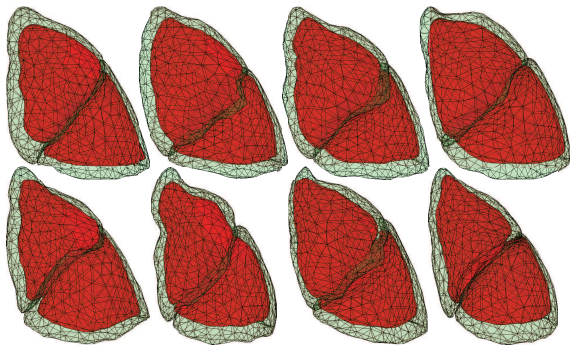}
\caption{The left lungs of live beagle dogs reconstructed as meshes from the acquired CT data. 
Pale green shows the inflated lung, and red shows the deflated lung. 
Ventral view.}
\label{fig:mesh_sample}
\end{figure}

The following procedure of data preprocessing was used.
First, the 3D surface mesh of lungs was reconstructed 
from acquired CT volume images. 
We employed the Synapse VINCENT image analysis system (Fujifilm Co., Ltd.) 
to perform anatomical segmentation of the upper and lower lobes 
from acquired CT volume images. 
Subsequently, each lobe was described by a mesh of points distributed throughout the surface of the lung. 
The surfaces of the lobes were generated as triangulated mesh representations 
using Poisson surface reconstruction \citep{Kazhdan2006}. 
The meshes of the two lobes were created independently, 
and each triangulated mesh was stored in the PLY file format.

Second, shape matching between inflated and deflated lungs was performed 
on the triangulated meshes.
The shape matching is required only for known lungs that contains the deflated shapes to make the training dataset. 
In applying the framework of this study to actual surgery, this procedure is not required in the estimation process of an unknown new lung. The correspondence of the position of the landmarks measured in deflated and inflated lungs were reconstructed from preoperative CT depending on the anatomical positions of landmarks.
On the process of shape matching, the number of vertices 
and the mesh topology corresponded between surfaces 
in the inflated and deflated states. 
The number of vertices in each lobe was set to 400. 
For the shape matching with high accuracy, 
we performed Laplacian-based surface registration using a differential displacement field. 
We employed the shape matching method in the previous study \citep{Nakao2019_CARS},
in which they analyzed surface deformation using deformable 
mesh registration and reported the shape matching method 
for inflated and deflated lung data. 
Our method estimates deformations 
for new cases by learning deformation from the registered model 
obtained by shape matching in \cite{Nakao2019_CARS}. 
Surface reconstruction and shape matching methods are not considered in this study, 
mainly to focus on the deformations due to deflation. 

The CT data of 11 lungs were taken at the same bronchial pressures, 
but each of them has different volume and volume ratio between inflated and deflated states.
The volumes and the volume ratios for each case are listed 
in Table 1 in \cite{Nakao2019_CARS}.
Using nine cases in which the volume was less than 60\% 
in both the upper and lower lobes, 
the volume ratio of the deflated lungs was unified to 60\%.
Some examples of the shape of the inflated and deflated lungs are shown in Figure \ref{fig:mesh_sample}.

As a data augmentation, the interpolated cases between the two original cases were created as augmented data.
The augmented data consisted of the midpoint of the vertices of the two meshes. 
Interpolated cases made from the original cases 
used for testing were not used for training. 
Only the original cases were used for testing, 
and no interpolated cases were used.
When the number of original cases used for training was $c$, the number of interpolated cases was $_c C _2$.

\subsection{\label{sec:Landmark_selection}Setting of few anatomical landmarks}
\begin{figure}[tb]
\includegraphics[width=85mm]{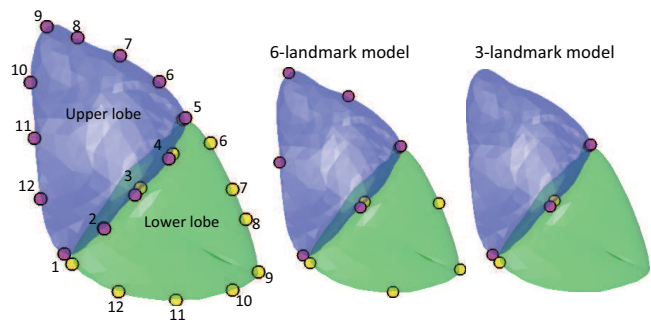}
\caption{Position of the landmarks. 
The pink and yellow markers indicate landmarks 
set in the upper and lower lobes, respectively. 
Landmarks were set separately on the upper and lower lobes. 
Landmarks 1--5 were placed 
on the side of the major fissure. 
The 3-landmark model and the 6-landmark model were set up as particularly few landmarks.}
\label{fig:landmark_order}
\end{figure}
The main goal of this study was to predict the deformation of the lung in the pneumothorax state from the lung in the inflated state using the positional information 
of a small number of landmarks.
The positions of landmarks were determined 
based on the anatomical shape of the whole lung, separately in the upper and lower lobes. 
Landmarks were placed at locations where it would be easy to measure the surface position during surgery.
It is difficult to set landmarks on the flat ventral surface because the surface of the lung has no anatomical pattern.
Therefore, we set landmarks along the outer surface of the lung, assuming that the 3D pointing device is used to trace the outer contour of the lung.

We first set up three landmarks at the corners of each lobe.
Next, we placed landmarks at the midpoint between the landmarks in those corners.
Subsequently, landmarks were placed at the midpoint of each of those landmarks.
As a result, 12 landmarks were set up.
The positions of the landmarks are visualized in Figure \ref{fig:landmark_order}. 
The numbers in the figure represent the order of the landmarks 
in Experiment 1 in Section \ref{sec:Obs_rate}. 
Starting from the edge of the major fissure, landmarks were placed around the left in the upper lobe and around the right in the lower lobe.
Landmarks 1--5 were located on the major fissure side.
For clinical applications, it is desirable to estimate the deformation from a small number of landmarks in a small area.
To validate the performance when the number of landmarks is limited, we used the 3-landmark model for landmarks 1, 3, and 5 out of 12 landmarks and the 6-landmark model for landmarks 1, 3, 5, 7, 9, and 11, as shown in Figure \ref{fig:landmark_order}.
To keep the area to be observed small, 
the landmarks were selected only from the side of the major fissure in the 3-landmark model.
In Experiment 2 of Section \ref{sec:Obs_rate}, we increased the landmarks in the order of numbers 1, 5, 3, 9, 7, 11, 2, 4, 6, 8, 10, and 12.
The third and sixth in Experiment 2 are the 3-landmark and 6-landmark models, respectively.

\subsection{\label{sec:data}Labeled dataset}
The problem of estimating deformations was solved using kernel regression. 
The regression was one of the supervised learning 
and required the use of a set of $D$ labeled samples, 
$\{ \bm{x}^{(d)}, \bm{y}^{(d)} \}_{d=1,2,...,D}$, 
where $\bm{x}^{(d)}$ was the $N$-dimensional input vector for the $d$th sample with an associated 
$M$-dimensional target output vector $\bm{y}^{(d)}$, 
which we intended to predict. 
We treated the features per vertex of the 3D mesh model as a single sample. 
In other words, $D$ is the total number of vertices in all mesh models used for training.
A data-driven method for estimating respiratory deformations 
of the liver using one vertex as a sample has been proposed by \cite{Lorente2017}. After preprocessing the acquired lung data, 
each mesh vertex in the deflated lung corresponded to a mesh vertex in the inflated lung with the same vertex number. 
Refer to the overview of the process in Figure \ref{fig:acq_proc}.
The black points, a blue point, blue arrows, and a yellow arrow represent landmarks, a target point, input vectors, and an output vector, respectively.

All vertices other than landmarks were used as target points $T$ to estimate deformations.
The following variables were available for each target point $T$ except for the landmarks: 
\begin{description}
  \item[$\bm{v}(T_{inf}), \bm{v}(T_{def})$] Coordinates of the vertex in the 3D space in the inflated and deflated lungs. 
  \item[$\bm{v}(L_{inf}^{(k)}), \bm{v}(L_{def}^{(k)})$] Coordinates of the $k$th landmark ($k = 1, 2, ..., l$) in the inflated and deflated lungs to which the vertex belonged.
  \item[$V_{inf}$] Whole volume of the inflated lung to which the vertex belonged.
  \item[$VR$] The volume ratio of the deflated lung to the inflated lung to which the vertex belonged.
\end{description}

Data for the upper and lower lobes, each consisting of 400 vertices, were created as separate datasets.
The number of samples $D$ used for training varied according to the number of cases $c$ and landmarks $l$; $D = (400-l)\times(c+ {_c C_2})$.
With all cases and six landmarks, the total number of samples was 14,184, except for one original case for the test. 
For each sample, $N = 3l\times2+2$ variables were used as inputs $\bm{x}$, including relative positional vectors 
$\bm{r}_{inf}^{(k)} = \bm{v}(T_{inf}) - \bm{v}(L_{inf}^{(k)})$ 
and $\bm{r}_{def}^{(k)} = \bm{v}(L_{def}^{(k)}) - \bm{v}(\overline{L_{def}})$, 
where $\overline{L_{def}}$ was the center point of all landmarks in deflated lung, 
$V_{inf}$ and $VR$. 
The target output vector $\bm{y}$ was the relative positional vector, 
$\bm{y} = \bm{v}(T_{def}) - \bm{v}(\overline{L_{def}})$,
then $M = 3$.

\subsection{\label{sec:Estimation}Kernel-based estimation of lung deformation}
In the kernel method, 
an output corresponding to a new input 
is estimated by solving the minimization problem, 
which is formulated as a ridge regression 
using the kernel matrix calculated from inputs of the labeled training dataset. 
Let $\bm{x}$ be the inputs of known dataset, 
we made the kernel matrix $K$ of Gaussian kernel. 
\begin{eqnarray}
\label{eq:kernel}
 K_{d,d'} & = & K (\bm{x}^{(d)},\bm{x}^{(d')}) \\
& = & k_a \exp \left\{-k_b \sum_{n=1}^N (x_n^{(d)}-x_n^{(d')})^2 \right\}.
\end{eqnarray}
Indices $d$ and $d'$ denote two different numbers of sample in the known training dataset
and $n$ is a component number of the input vector. 
Parameters $k_a$ and $k_a$ were determined by cross-validation. 

The problem to minimize the squared error with the output $\bm{y}$ of training data 
was formulated using the calculated kernel matrix $K$ 
with L$_2$ norm regularization as 
\begin{eqnarray}
\label{eq:minfunc}
 \min_{\bm{W}} \left\{ \frac{1}{2} \sum_d^D (\bm{y}^{(d)} - \sum_{d'}^D K_{d,d'} W_{d'})^2 + \frac{\lambda}{2} \sum_{d,d'}^D W_{d} K_{d,d'} W_{d'} \right\}. \nonumber \\ 
\end{eqnarray}
The regularization parameter $\lambda$ was determined by trial and error. 
The solution of $\bm{W^*}$ was 
\begin{eqnarray}
\label{eq:W}
\bm{W^*} = (K + \lambda E)^{-1} \bm{y}, 
\end{eqnarray}
where $E$ is an identity matrix of the same size as the kernel matrix $K$. 

Subsequently, an unknown output $\bm{y}^{(i)}$ corresponding to a new input $\bm{x}^{(i)}$ 
for a test data $i$ was estimated as 
\begin{eqnarray}
\label{eq:predict}
 \bm{y}^{(i)} = \sum_{d}^D K_{i,d} W_{d}, 
\end{eqnarray}
where $K_{i,d} = K (\bm{x}^{(i)},\bm{x}^{(d)})$.

\subsection{Experimental settings}
We tested the estimation performance when using 6 and 3-landmark models, when using different numbers of landmarks, and when using different numbers of cases for training.
In all experiments, 
we performed cross-validation using all cases as test data once.
The experimental results with the smallest estimation error are described.
We compared the results of our method with 
those of affine transformation (AF) and thin-plate spline method (TPS) 
to evaluate the performance of our developed method. 
In the previous studies that estimated the lung shape in the pneumothorax state, 
AF was the method used for lung surface registration 
during the registration of the internal structure of the lung \citep{Uneri2013}, 
and TPS was the method for non-rigid registration \citep{Nakamoto2006}.

\section{\label{sec:Exp_res}Results}
\subsection{\label{sec:Est_res}Estimation results}

\begin{figure}[tb]
\includegraphics[width=80mm]{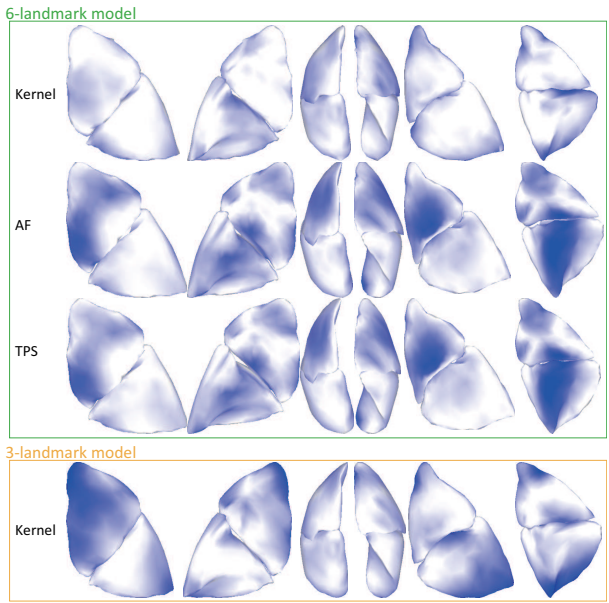}
\caption{Estimated shape of the deflated lungs 
using the proposed method (Kernel), affine transformation (AF), and thin-plate spline method (TPS). 
The magnitude of the local positional error is shown in blue with a maximum of 8.5 mm. 
Each image shows a view from the same direction of the results 
of testing the same cases.}
\label{fig:Estimation_result}
\end{figure}

\begin{table}[tb]
\caption{\label{tab:Estimation_result}
Average (standard deviation) values of the deformation estimation results for RMSE, Dice similarity coefficient (DSC), and Hausdorff distance (HD).}
\begin{ruledtabular}
\begin{tabular}{llll}
 &\multicolumn{3}{l}{6-landmark model}\\
\cline{2-4}
Metric &	Kernel	&	AF	&	TPS	\\
\hline
Upper lobe &&&\\
RMSE [mm]    &2.74 (0.52)&3.58 (0.81)&3.44 (0.83)		\\
DSC (0--1)     &0.90 (0.02)&0.80 (0.06)&0.81 (0.06)		\\	
HD [mm]      &6.11 (1.28)&7.65 (1.79)&7.63 (1.88)		\\	
\hline
Lower lobe &&&\\
RMSE [mm]    &3.04 (0.63)&3.56 (0.55)&3.40 (0.51)		\\	
DSC (0--1)     &0.94 (0.01)&0.90 (0.03)&0.90 (0.03)		\\	
HD [mm]      &6.29 (1.11)&8.38 (1.43)&8.45 (1.54)		\\
\end{tabular}
\begin{tabular}{llll}
 &\multicolumn{3}{l}{3-landmark model}\\
\cline{2-4}
Metric &	Kernel	&	AF	&	TPS	\\
\hline
Upper lobe &&&\\
RMSE [mm]    &3.86 (0.90)&40.79 (5.26)&40.79 (5.26)		\\
DSC (0--1)     &0.87 (0.04)&0.00 (0.00)&0.00 (0.00)		\\	
HD [mm]      &7.87 (0.74)&85.38 (10.77)&85.38 (10.77)	\\	
\hline
Lower lobe &&&\\
RMSE [mm]    &3.55 (0.85)&44.52 (5.99)&44.52 (5.99)		\\	
DSC (0--1)     &0.91 (0.02)&0.00 (0.00)&0.00 (0.00)		\\	
HD [mm]      &7.41 (2.15)&94.81 (11.98)&94.81 (11.98)	\\
\end{tabular}
\end{ruledtabular}
\end{table}

First, we present the estimation results using the 6-landmark model and 3-landmark model. 
Figure \ref{fig:Estimation_result} 
visualizes some estimated results using 6 and 3-landmark model. 
The shape shown in the figure is the estimated lung in the pneumothorax state. 
The magnitude of the local positional error is shown in blue with a maximum of 8.5 mm.
In the 3-landmark model, the results estimated using AF and TPS did not result in a lung shape.

To quantitatively evaluate the results of deformation estimation, 
we calculated the root mean square error (RMSE), the Dice similarity coefficient (DSC), and the Hausdorff distance (HD). 
The RMSE was the root mean squared error of all vertices of the local positional error per vertex. 
The DSC was calculated as the percentage of overlap 
between the 3D meshes of the estimated and correct deflated lungs. 
Hausdorff distances were calculated by considering 
the vertex group of the mesh as a point group. 
The RMSE and Hausdorff distances are the distance-based estimators in millimeters, 
and the smaller the value, the better the estimate. 
The DSC as the volume-based estimator is 0 for the minimum and 1 for the maximum, 
and the larger the value, the better the estimation result.
The average and standard deviation of the RMSE, DSC, 
and HD for the upper and lower lobes 
for the proposed method using the kernel, AF, and TPS 
are summarized in Table \ref{tab:Estimation_result}.

\subsection{\label{sec:Obs_rate}Observation rate}

\begin{figure*}[tb]
\includegraphics[width=180mm]{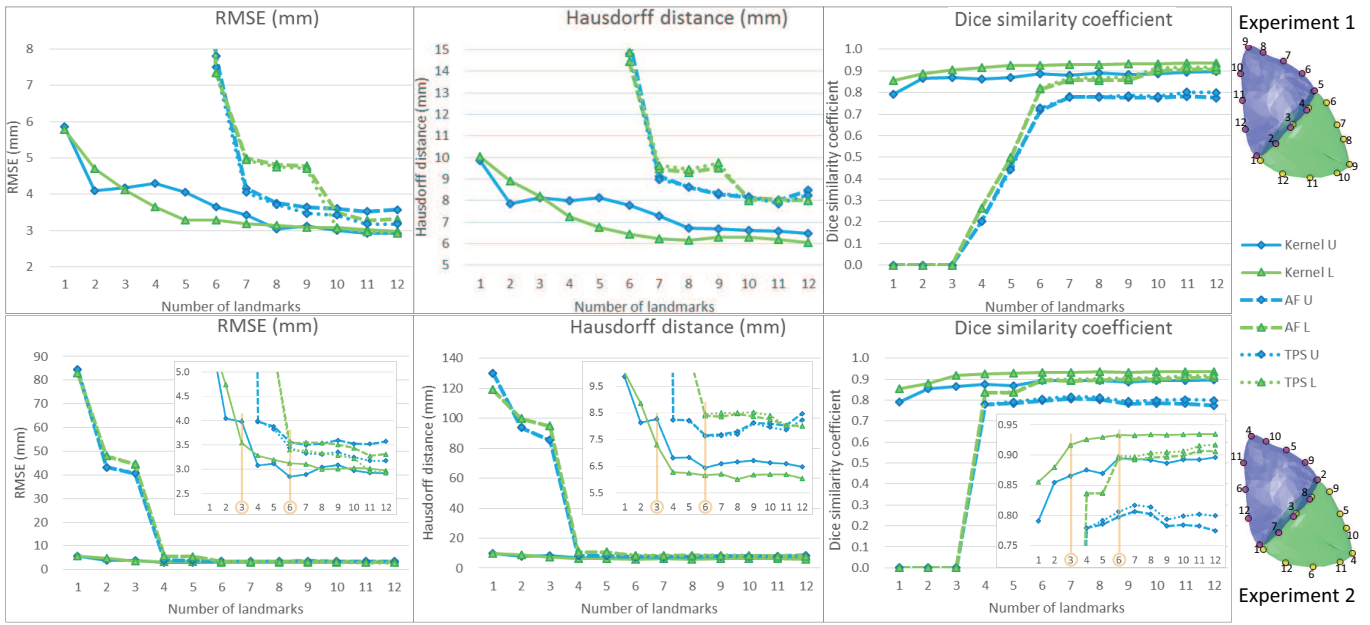}
\caption{Estimated results for a different number of landmarks. 
The numbers on the lung model indicate the order of the landmarks in each experiment.
The top row shows the results of Experiment 1 when landmarks were increased in sequence around each lobe, and the bottom row shows the results of Experiment 2 when landmarks were increased to include 3 and 6-landmark models (orange marker).
The blue line with diamond-shaped markers represents the upper lobe (U), 
and the green line with triangular markers represents the lower lobe (L) results. 
The solid, dashed, and dotted lines show the results of 
the proposed method (Kernel), AF, and TPS, respectively.
The small figures in the bottom figures are enlarged versions of the small error area. }
\label{fig:Result_land}
\end{figure*}

The relationship between the number of landmarks to be observed 
and the estimation accuracy was investigated.
To examine the area of the lungs covered by the landmarks, we conducted two experiments, in which we changed the order of the landmarks.
The landmarks were increased in the order of the number shown on the lung in Figure \ref{fig:Result_land}.
In Experiment 1, 
landmarks were increased around the left in the upper lobe and around the right in the lower lobe, tracing the outer contour of the lung. 
In Experiment 2, 
landmarks 1--6 were selected sparsely throughout the outer contour of the lung, and landmarks 7--12 were selected to fill the space between them.

Figure \ref{fig:Result_land} shows the RMSE, Hausdorff distance, and Dice similarity coefficient of the estimation results 
using our method (Kernel), AF, and TPS. 
The solid, dashed, and dotted lines show the results of the Kernel, AF, 
and TPS, respectively. 
In the graphs of Experiment 2, 
the points marked in orange line are the results when 3- and 6-landmark models are used.
The smaller figures in the figures are enlarged graphs of the area with small values on the vertical axis.

\subsection{\label{sec:Data_var}Case variation}
The estimated performance was confirmed with 3 and 6-landmark models 
by varying the number of cases used for training. 
We experimented with all combinations of cases to exclude and a case to test.
Augmented data interpolated between the two cases were included 
only for the cases used for training.

\begin{figure*}[tb]
\includegraphics[width=180mm]{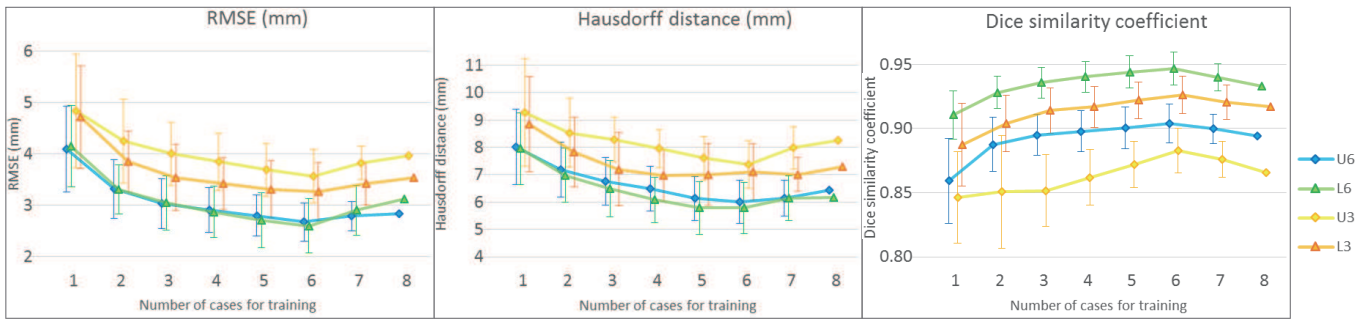}
\caption{Results depending on the number of cases 
included in the training (one to eight).  
A horizontal axis of 8 is when all the cases were used 
for training except for the test. 
Blue and green represent the results of 6-landmark model (6), 
while yellow and orange represent the results of 3-landmark model (3). 
The diamond-shaped markers indicate the upper lobe (U), 
and the triangular markers indicate the lower lobe (L). 
The standard deviation of the results for different cases 
included in the training is added as an error bar.}
\label{fig:Result_case}
\end{figure*}

Figure \ref{fig:Result_case} shows the results 
when the number of cases included in the training was from one to eight. 
The average value of the results for all combinations 
that changed the excluded cases is shown as a line graph, 
with the standard deviation as an error bar. 

\section{\label{sec:Disc_Conc}Discussion and Conclusions}
We proposed a method for estimating 
the overall intraoperative lung surface profile 
from the partial intraoperative surface information and the preoperative lung surface profile using kernel regression. 
To the best of our knowledge, 
this is the first study to show that lung deformation 
in the pneumothorax state is predicted 
from very few landmark positions in a data-driven approach. 
Previous studies have focused on the registration of the internal structures of the lung, 
using intraoperative CBCT or estimating lung deformations with mechanical models.
On the other hand, this study employed the surface deformations of the lung 
and proposed a method that required fewer measurements intraoperatively. 
Indeed as shown in Section \ref{sec:Est_res}, 
we listed the results of estimating the pneumothorax deformation 
from three or six landmarks placed on the surface of the lung 
using our method, AF, and TPS. 
In the 6-landmark model, the estimation errors evaluated by the RMSE, DSC, and HD 
when using our method were all smaller than the estimation errors 
when using the two existing methods for both of upper and lower lobes.
The location of the landmarks in the 6-landmark model covers three sides of outer contour of the lung surface.
In the 3-landmark model, our method estimated the deformation with the RMSE 
that was only about 1 mm larger than the RMSE in 6-landmark model.
The AF and TPS did not provide sufficient estimation results when using the 3-landmark model.
The landmarks in the 3-landmark model are located only in the major fissure side.

As shown in Section \ref{sec:Obs_rate}, 
our method fairly well estimated the deflated lung surface with very few measurement points
within a limited area in the major fissure side. 
On the other hand, although the two conventional methods 
could estimate the deformation with a relatively small error 
when landmarks were placed on two or more sides of each lobe, 
they did not provide sufficient results 
when the landmarks were selected only from the major fissure side. 
Estimation accuracy was much better when landmarks in the region outside the major fissure side were included.
The results of Experiment 2 showed that the error did not change much between the 6-landmark model, in which half of the 12 landmarks were selected for one skip, and when all 12 landmarks were selected.
The results of five landmarks in Experiment 1 and three landmarks in Experiment 2, where landmarks were placed only on the major fissure side, did not change much in the estimation accuracy of our method.
The observed coverage area of the lung affected the estimation performance, but the number of landmarks had little effect.
Nonetheless, intraoperative measurement with a 3D position sensor would increase the physician's burden.
To solve this problem, a technique to measure the lung surface 
with a laser pointer for the registration has been developed by \cite{Nakamoto2007}. 

Section \ref{sec:Data_var} reported the estimation performance of our method 
when the number of cases to be trained was reduced.
Although a large amount of data would be needed for a data-driven estimation, 
it is difficult to obtain a large number of case data for clinical applications.
The maximum number of cases for training in this study was eight. 
The average of the RMSE and DSC had the smallest test error 
when the number of cases was six in both the upper and lower lobes.
The error dropped significantly from one to three cases, 
but not much after four cases. 
Three cases were enough for the training in our method to produce results with a sufficiently small error.
The kernel method enables to estimate deformation with small errors 
even with a small number of samples. 
Therefore, it is suitable for clinical applications 
where burden of data acquisition is high.

A technical limitation of our method is 
that it relies on the method of shape matching of inflated 
and deflated shapes of known lungs to build the training dataset.
The shape matching method is considered to be outside the scope of this study. 
Registration errors that occur during shape matching 
would be included in the learning by our method.
As a recent study of shape matching, 
regularized keypoint matching improves deformable registration 
in lung CT \citep{Ruhaak2017}. 
Training with more accurate shape-matched data 
may reduce the error in the deformation estimation by our method. 

In addition, measurement errors when measuring the position of landmarks 
during surgery should be considered.
Then we investigated the effect of measurement error of landmark positions 
on the estimation, both analytically and numerically.
We perform the first-order Taylor expansion of the kernel matrix $K_{i,d}$ 
on the right-hand side in Eq.(\ref{eq:predict}) 
and attain a relation $\Delta \bm{y} = A \Delta \bm{x}$, 
where $\Delta \bm{x}$ represents the measurement error and 
its effect is denoted by $\Delta \bm{y}$. 
This equality demonstrates magnitude of $\Delta \bm{y}$ due to the measurement error $\Delta \bm{x}$.
For simplicity, we compute the eigenvalues of the matrix $A^T A$ (singulars value of $A$) 
and assess the empirical mean and variance of the maximum 
of the eigenvalues for the whole experimental dataset for each lobe 
as $\Lambda = 0.232 \pm 0.063$ for the upper lobe 
and $\Lambda = 0.245 \pm 0.037$ for the lower lobe. 
In other words, the effects from measurement error becomes about 
$\Delta \bm{y} \approx \sqrt{0.232} \times \Delta \bm{x}$ or 
$\Delta \bm{y} \approx \sqrt{0.245} \times \Delta \bm{x}$.

In the training procedure, 
a single vertex deformation was trained as a single sample  
in a 3D mesh constructed on inflated and deflated lungs.
The reason for using one vertex as one sample  
rather than the entire lung mesh as one sample is 
that we considered including lungs with different numbers of mesh vertices 
in the training in the same framework.
All the acquired cases were processed at one time, 
and the number of mesh vertices and topology for all cases was the same.
However, in future clinical applications, 
the structure of the meshes constructed from newly acquired CT data 
would be different.
Then our method does not need to change the training framework and the landmark setting procedure 
even if the structure of the meshes in each case is different.

In future work, we consider a clinical application 
of our deformation estimation method. 
In our experiment, 
we employed the CT images of the left lungs of live beagle dogs.
One direction of the future studies is 
investigating the estimation performance of our method 
when applied to the right lung and the lung of human.
Since the lungs of dog and human are different in size, 
the kernel matrix learned for lungs of dog cannot be directly 
used to estimate the deformation of the lung of human.
Therefore, when our method is applied to human, 
it is necessary to re-train the kernel matrix 
with the lung data of human. 
Nevertheless the training requires a short time 
because the training method in this study is 
very computationally inexpensive.
In the human case, CT images of the lung in the inflated state 
can be obtained preoperatively. 
The CT image of the lung in the pneumothorax state is supposed to be 
reconstructed from the CBCT image during the surgery.
In addition the framework of the kernel regression can be used for intraoperative CBCT imaging \citep{Nakao2020}.
The findings of this study would be useful for the deformable registration of CBCT images, 
which have issues such as low contrast points and motion artifacts. 
By identifying a few landmarks shown in this study, 
it is expected to achieve highly accurate registration of CBCT images.
With the addition of deformation information to the real-time CBCT images 
used in daily surgery, 
it would be possible to realize more efficient surgical navigation.
Furthermore, data-driven estimation of pneumothorax deformations, 
which are difficult to represent with mechanical models, 
also contribute to the statistical modeling of pneumothorax deformations.

\begin{acknowledgments}
This research was supported by Japan Agency for Medical Research and Development (AMED) and Acceleration Transformative Research for Medical Innovation (ACTM) Program. 
A part of this study was also supported by 
Japan Society for the Promotion of Science (JSPS) Grant-in-Aid for Young Scientists (B) [grant number 16K16407], 
JSPS Grant-in-Aid for Early-Career Scientists [grant number 19K20709], and 
Grant-in-Aid for JSPS Fellows [grant number 20J40290].
\end{acknowledgments}

\bibliographystyle{apa.bst}
\bibliography{Bib_deform_deairation}

\end{document}